\documentclass[sigconf]{acmart}

\renewcommand\footnotetextcopyrightpermission[1]{} 
\usepackage{booktabs} 

\newcommand{\MODEL}{{\it WESPAD }}
\newcommand{\PROBLEM}{{\it PHM }} 
\newcommand{\eg}{{e.g., }} 

\usepackage{enumitem}
\usepackage{balance}

\begin{document}

\copyrightyear{2018}
\acmYear{2018} 
\setcopyright{iw3c2w3}
\acmConference[WWW 2018]{The 2018 Web Conference}{April 23--27, 2018}{Lyon, France}
\acmPrice{}
\acmDOI{10.1145/3178876.3186055}
\acmISBN{978-1-4503-5639-8/18/04}
\fancyhead{}

\title{Did You Really Just Have a Heart Attack? Towards Robust Detection of Personal Health Mentions in Social Media}


\author{Payam Karisani}
\affiliation{%
  \institution{Emory University}
}
\email{payam.karisani@emory.edu}

\author{Eugene Agichtein}
\affiliation{%
  \institution{Emory University}
}
\email{eugene.agichtein@emory.edu}

\begin{abstract}
Millions of users share their experiences on social media sites, such as Twitter, which in turn generate valuable data for public health monitoring, digital epidemiology, and other analyses of population health at global scale. The first, critical, task for these applications is classifying whether a personal health event was mentioned, which we call the (\PROBLEM) problem. This task is challenging for many reasons, including typically short length of social media posts, inventive spelling and lexicons, and figurative language, including hyperbole using diseases like ``heart attack'' or ``cancer'' for emphasis, and not as a health self-report. This problem is even more challenging for rarely reported, or frequent but ambiguously expressed conditions, such as ``stroke''. To address this problem, we propose a general, robust method for detecting PHMs in social media, which we call \MODEL, that combines lexical, syntactic, word embedding-based, and context-based features. \MODEL is able to generalize from few examples by automatically distorting the word embedding space to most effectively detect the true health mentions. Unlike previously proposed state-of-the-art supervised and deep-learning techniques, \MODEL requires relatively little training data, which makes it possible to adapt, with minimal effort, to each new disease and condition. We evaluate \MODEL on both an established publicly available Flu detection benchmark, and on a new dataset that we have constructed with mentions of multiple health conditions. Our experiments show that \MODEL outperforms the baselines and state-of-the-art methods, especially in cases when the number and proportion of true health mentions in the training data is small. 
\end{abstract}

%
%




\maketitle

\section{Introduction} 
\label{sec:intro}

Individuals and organizations increasingly rely on social data, created on platforms such as Twitter or Facebook, for sharing information or communicating with others. Large volumes of this data have been available for research, opening new opportunities to answer questions about society, language, human behavior, and health. Among these, monitoring and analyzing social data for public health has been an active area of research, due to both the importance of the topic, and to the unprecedented opportunities afforded by a real-time window into the self-reported experience of millions of people online. 
These social data come with many challenges and potential biases\cite{olteanu2016social}. Nevertheless, these data already enabled many public health applications, such as tracking the spread of influenza\cite{aramaki2011twitter,paul2011you}, understanding suicide ideation\cite{de2016suicide}, monitoring and providing support during humanitarian crises\cite{imran2016crisis}, drug use \cite{daniulaityte2016bad,prier2011identifying}, drinking problems \cite{moreno2012associations}, and public reactions to vaccination~\cite{salathe2011assessing}.

The main advantages of social data over traditional methods of public health surveillance such as phone surveys, in-person interviews, and clinical reports, include scalability and potential near-real time responsiveness. 
Therefore, social data has become a valuable source to monitor and analyze people's reports and reactions to health-related incidents. A crucial first step in disease analysis and surveillance using social data, is to identify whether a user post is actually mentioning a specific person reporting a health event. All subsequent processing and analysis, whether it is epidemic detection (\eg mentioning an affected person in the post), or individual analysis (\eg reporting one's own health condition), depends on the accuracy of the detection and categorization of the individual postings. If the posting were mis-categorized, and did not in fact report a health-related event, all subsequent analysis and conclusions arising from the data might be flawed.

Our goal is to accurately identify postings in social data, which not only contain a specific disease or condition, but also mention a person who is affected. For instance, we aim to identify posts such as: \textit{"My grandpa has Alzheimer's \& he keeps singing songs about wanting to forget"} or \textit{"Yo Anti-Smoking group that advertises on twitch, I don't smoke. My mom died to lung cancer thanks to smoking for like 40 years. I get it."}. In contrast, we wish to filter out postings like: \textit{"I almost had a heart attack when I found out they're doing a lettering workshop at @heathceramics in SF"} or \textit{"Dani seems like a cancer, spreads herself anywhere for attention!"}. In terms of previous work, we aim to identify specific health reports, rather than non-relevant postings or postings expressing general concern or awareness of a disease or condition \cite{lamb2013separating}. We call this task detecting Personal Health Mentions, or \PROBLEM. Further, we aim to develop a solution that is both robust and general, so that it can scale to many diseases or conditions of current or future interest. In turn, more accurately detecting personal health mentions in social data, without requiring extensive disease-specific development and tuning, would empower public health researchers, digital epidemiologists and computational social scientists to ask new questions, and to answer them with higher confidence.


Detecting health mentions in social data is a challenging task. Social data posts, such as those on Twitter, tend to be short, and are often written informally, using diverse dialects, and inventive and specialized lexicons. Previous efforts for similar tasks applied machine learning methods that relied on extensive feature engineering, or on external feature augmentation to address the sparsity in the feature space, e.g., for company name detection \cite{spina2013discovering}, reputation measurement \citep{cha2010measuring}, sarcasm detection \cite{bamman2015contextualized,joshi2016word}, and for public health \cite{lamb2013separating,de2013predicting}.
In the health context, the problem is exacerbated by the limited availability of training data, and by the low frequency of the health reports in even keyword-based samples: on Twitter, our experiments show that of the tweets containing a disease name keyword, only 19\% are actual health reports. 
The resulting classifiers tend to have high precision, but relatively low recall (i.e., high false negative rate), which may not be desirable for applications such as disease surveillance or detecting epidemics. 


Our goal is to address the problems of sparsity and imbalanced training data for \PROBLEM detection, by explicitly modeling the differences in the distribution of the training examples in the word embeddings space. For this, we introduce a novel social text data classification method, \MODEL (\textbf{W}ord \textbf{E}mbedding \textbf{S}pace \textbf{Pa}rtitioning and \textbf{D}istortion), which learns to {\em partition} the word embeddings space to more effectively generalize from few training examples, and to {\em distort} the embeddings space to more effectively separate examples of true health mentions from the rest. 
While deep neural networks have been proposed for this purpose, our method works well even with small amounts of training data, and is more simple and intuitive to tune for each new task, as we show empirically in this paper. 
We emphasize that \MODEL requires no topic or disease-specific feature engineering, and fewer training examples than previously reported methods, while more accurately detecting true health mentions. 
These properties make \MODEL particularly valuable for extending public health monitoring to a wider range of diseases and conditions. Specifically, our contributions are:

\begin{itemize}[leftmargin=0.6cm]
\item We propose a novel, general approach to discover domain-specific signals in word embeddings to overcome the challenges inherent in the \PROBLEM problem. 
\item We implement our approach as an effective \PROBLEM classification method, which outperforms the state-of-the-art classifiers in the majority of settings.
\item To validate our approach, we have constructed and released a manually-annotated dataset of Twitter posts for six prominent diseases and conditions\footnote{The dataset and code are available at~\url{ https://github.com/emory-irlab/PHM2017}}.
\end{itemize}
Next, we review related work to place our contributions in context.

\section{Related work} 
\label{sec:related}



Social network data and user-contributed posts on platforms such as Facebook and Twitter, have been extensively studied for diverse applications in business, politics, science, and public health. Some prominent examples include work on 
answering social science questions \cite{lazer2009life}, and analyzing influenza epidemics \cite{chew2010pandemics}. Our work builds on three general directions in this area: general classification techniques that serve as the foundation of our work; disease-specific classifiers for social text data; and, closest to our work, prior research on general health classifiers that could be potentially applied to different diseases and conditions.

\subsection{Text Classification: Models and Techniques}
Methods for automatic text classification have been studied for decades, and have evolved from simple bag-of-words models to sophisticated algorithms incorporating lexical, syntactic, and semantic information \cite{aggarwal2012survey}. Many of these algorithms have been adapted for biomedical text processing, with varying success \cite{cohen2005survey,paul2017social}.
Recently, deep neural networks have emerged in many areas of natural language processing as an alternative to feature engineering and demonstrating new state-of-the-art performance on a wide range of tasks. However, there are two main challenges in making these models effective. First, it is commonly known that deep neural models usually need a large amount of training data to reach their ultimate capacity. This is an active area of research, and workshops such as Limited Labeled Data (LLD)\footnote{Available at~\url{https://lld-workshop.github.io/}} are held to investigate this area. Second, it is not clear how to incorporate domain knowledge into the training--\eg social network topology, or user activities. Nevertheless, we include three state-of-the-art deep neural network models as baselines, chosen as representative of the most effective neural network methods reported for text classification. We show empirically that our proposed method performs better than these techniques, especially in the settings with small amounts of available training data.

To improve the generalization of classification to unseen textual cases, {\em word embeddings} \cite{mikolov2013distributed} have been proposed as a semantic, and broader, representation of text. This idea has been used, for example, to compare two sentences, in addition to lexical features. For instance, \cite{mihalcea2014simcompass} proposed a system for detecting semantic similarity between two pieces of texts using word embeddings similarity. 
\cite{kenter2015short} proposed an algorithm similar to \cite{mihalcea2014simcompass}, for paraphrase detection task. Their main contribution is that the algorithm can capture the similarity between two sentences with higher details through binning the word similarity values. For another task, sarcasm detection, reference~\cite{joshi2016word} evaluated a number of word embeddings features to discover word incongruity, by measuring the similarities between the word vectors in the sentence. While these studies have been done for different domains and tasks, they all share the same idea of using word embeddings space as a resource to extract features; we also build on this general idea for the \PROBLEM detection problem. 
Reference~\cite{yu2013compound} proposed the idea of clusters of word vectors to address the problems of word ambiguity. The clusters are used to generate compound embeddings features. In our experiments we observed that word clusters do not accurately characterize social data texts when used directly. Instead, we propose {\em partitioning} the word embeddings space to generate features describing the distribution of training examples in the different regions of the space. The resulting distributions are subsequently used to map the instances of each class to different categories, which, as we show allows \MODEL to more precisely identify true instances of \PROBLEM. 
\vspace{-0.4cm}

\subsection{Text Classification for Health}

\noindent\textbf{Disease-specific text classifiers}:
Since building a large training set for public health monitoring is costly, and in some cases  impossible (e.g., for rarely reported diseases); it has been shown that domain knowledge in the form of rule-based, or domain-specific classifiers is effective in monitoring certain diseases, e.g., \cite{lamb2013separating,de2013predicting}. A large body of work has been done for detecting and tracking information about specific diseases. This includes investigations on tracking the spread of flu epidemics \cite{aramaki2011twitter,lamb2013separating}, cancer analysis \cite{ofran2012patterns}, asthma prediction \cite{dai2017predicting}, depression prediction \cite{de2013predicting,yazdavar2017semi}, and anorexia characterization \cite{de2015anorexia}. To improve accuracy for each of these domains, studies such as \cite{lamb2013separating} have shown that certain aspects of tweets are also good indicators of health reports, and have been successfully operationalized as lexical and syntactic features, which we incorporate into our baseline system. A thorough overview of the published papers can be found in references \cite{charles2015using,paul2017social}. Our work can be potentially used to improve the accuracy of health mention detection for {\em all} of the mentioned disease specific studies. We emphasize that an advantage of our model, introduced in the next section, is that without imposing any restriction on the original features, our method can substantially improve the detection accuracy, even when there is only a small set of positive examples available.\\
\noindent\textbf{General-purpose text classification for Health}
A more attractive strategy than developing disease-specific classifiers, is to develop a single classification algorithm, that could be easily adapted to detect mentions of different diseases and conditions. This is the direction we chose in this work. 
Reference \cite{paul2011you} reports using an LDA topic-based model, which also incorporates domain knowledge, to discover the symptoms and their associated ailments in Twitter. \cite{prieto2014twitter} proposed a two step process, which is representative of a common methodology, to detect the health mentions in social text data. The first, high-recall step is to collect the tweets using keywords and regular expressions, and the second step is to use a high-precision classifier -- in this case, by using a correlation-based feature extraction method. Reference~\cite{yin2015scalable} reported using a dataset of tweets across 34 health topics, and investigated the accuracy of the classifiers trained over multiple diseases and tested on new diseases. The authors conclude that training a classifier on four diseases: cancer, depression, hypertension, and leukemia can lead to a general health classifier with 77\% accuracy using standard SVM classifiers and bag-of-words features, similar to one of our baselines used in empirical evaluation, which, as we will show, is not able to generalize well to unseen test data.
We emphasize that our aim is also to develop a general health mention detection model that could apply to a variety of diseases and conditions. 

In summary, to our knowledge, our \MODEL model (presented next) is the first {\em general} health report detection method that requires only small amounts of training data, does not do any domain-specific feature engineering, yet performs as well as, and often better than, other methods, including a disease-specific rule-based classifier.

\section{\MODEL Model Description} 
\label{sec:model}
This section introduces our method, \MODEL, for robust classification of health mentions in social data. We first summarize previously proposed lexical and syntactic features for social media classification, used both for health mentions, and other domains, which we use as starting point for our method. We then introduce the novel steps of our work, for learning topic-specific representation of the data derived from word embeddings (Sections~\ref{subsec:par-emb} and \ref{subsec:we-mean}). 

\subsection{Lexical and Syntactical Features} \label{subsec:lex-syn}
Previous studies on analyzing social media (primarily Twitter and Facebook posts) for depression prediction \cite{de2013predicting}, influenza tracking \cite{lamb2013separating}, and tobacco use \cite{prier2011identifying}, have shown that certain words and phrases are key indicators of the health reports. Therefore, we use all word unigrams and bigrams as features, in order to capture any  words or phrases that may be salient.

Additionally, to model syntactic dependencies in the text, we use the approach proposed in \cite{matsumoto2005sentiment} to identify common syntactical dependencies in the tweets through detecting the frequent syntactical subtrees, and use them as features. To detect the sentences in the posts, we used the tweet dependency trees \cite{kong2014dependency} to detect sentence boundaries.
We conjecture that even with small amount of training data, the frequent subtrees can automatically detect a subset of syntactic patterns which are usually designed manually for certain health cases (such as those of flu detection in \cite{lamb2013separating}). Our experiments (in Section \ref{sec:results}) show that lexical and syntactic features provide high precision for health mention detection, but are not sufficient to generalize from the (relatively small) amounts of training data. To improve generalization, we now describe our use of word embeddings, which allows learning from a few positive examples of health mentions. In the next sections, we use word \textit{lex\_feats} to refer to the bigrams, and use word \textit{syn\_feats} to refer to the features extracted from the frequent subtrees.

\subsection{Detecting ``Noisy'' Regions in the Word Embeddings Space} 
\label{subsec:new-mean}

Word embeddings \cite{mikolov2013distributed} is an approach to map words to linguistic concepts in a lower-dimensional vector space. The motivation for using word embeddings to address sparsity in our task is that it could help match and generalize training examples to unseen examples at test time, which may not share the same words but have semantically related meaning. 
A common way to represent a short piece of text in the word embeddings space is to average of the constituent word vectors, and use the centroid of the vectors directly as features for a classifier. Although this approach has some drawbacks, \eg losing information about individual words, several studies have shown that it can be effective \cite{kenter2015short,mihalcea2014simcompass,socher2013reasoning}. Here we explore the ways that the centroid representation can be extended to improve classifier generalization to unseen cases.

One could incorporate the classifier output (predicted class) alongside other features in the final feature vector. However, as we will show, incorporating the classifier output, as is, would propagate the false positive matches in the word embeddings space to generate noisy features for the final classifier.
Another problem with the approach above is that a centroid in the word embeddings space does not preserve information about the constituent words, and thus likely to map both positive and negative examples of a health mention (if they share common words) into similar vectors in the word embeddings space.
However, we will use the centroids as a starting point, in combination with a classifier trained over the centroid features, to detect, and downweight, the regions in the word embeddings space, where the positive and negative training examples have such a similar centroid vectors that they are no longer distinguishable.


\noindent\textbf{Definition: ``Noisy'' regions in the word embeddings space}: We define ``noisy'' regions as those, where the precision of a centroid-based classifier is lower than a certain threshold $\alpha$. 

Using this definition, we can now {\em filter} out the noisy regions (and the corresponding features from training data). Detecting the regions which are {\em not} noisy also can help us to address the challenge of an imbalanced training set. Examples with centroids mapped to these regions can be directly used in the model to predict the label of the instances which are not present in the training set. This can lead to a model which can generalize better with only a small set of positive cases. To detect the noisy regions in the embeddings space we define a probabilistic function \(Pr\) to be the probability of assigning the tweet centroid to the positive class. The values associated with \(Pr\) can be extracted from the training set using the logistic regression model as the centroid-based classifier. Given the function \(Pr\) and the associated values, the probability of assigning tweet \(t_i\) to the positive class would be \(Pr(t_i)\). Based on the definition, in the noisy regions the value of \(Pr\) is close to 0.5. More formally, we define binary features \(PFlag(t_i)\) and \(NFlag(t_i)\) for tweet \(t_i\) as follows:

\begin{displaymath}
	PFlag(t_i) = \begin{cases}
	1 & 0.5 + \alpha \leq Pr(t_i)\\
	0 & Otherwise
	\end{cases}
\end{displaymath}

\begin{displaymath}
	NFlag(t_i) = \begin{cases}
	1 & 0.5 - \alpha \geq Pr(t_i)\\
	0 & Otherwise
	\end{cases}
\end{displaymath}
in which $\alpha$ is the threshold to detect the noisy regions, and can be tuned in the training phase. If tweet \(t_i\) is predicted to be positive and is not located in a noisy region, the value of \(PFlag(t_i)\) is set, and likewise, if it is predicted to be negative, and is not located in the noisy regions the value of \(NFlag(t_i)\) is set. The output of one example of the noisy region detection is illustrated in Figure~\ref{fig:2d-space-combined}(a). Figure~\ref{fig:2d-space-combined}(a) illustrates a 2-dimensional projection of positive and negative examples (marked with 'x' and 'o', respectively) using t-SNE \cite{maaten2008visualizing}, and the corresponding noisy region, the circle area where the centroids of positive and negative examples are not distinguishable with high confidence. All the data points contain word \textit{heart~attack}.

\subsection{Partitioning the Word Embeddings Space} 
\label{subsec:par-emb}

\begin{figure*}
\includegraphics[height=2in, width=6.6in]{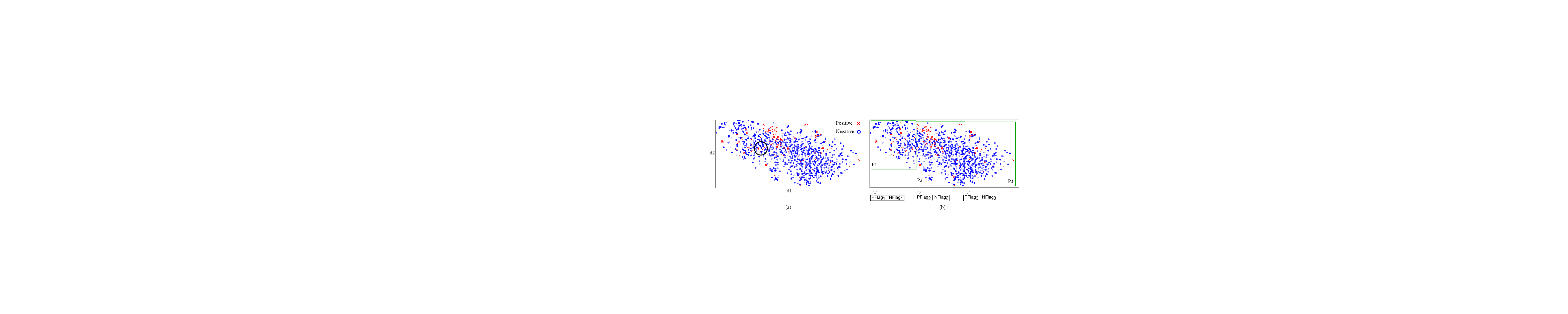}
\caption{ (a) Tweet centroids in the word embeddings space that contain the phrase \textit{heart attack}, projected to two dimensional space using t-SNE. (b) The same word embeddings space with 3 hypothetical partitions, and a pair of features associated with each partition.} 
\label{fig:2d-space-combined}
\vspace{-0.2cm}
\end{figure*}

The ``noisy region'' flags \(PFlag\) and \(NFlag\) can help us capture the semantic similarity between the tweets which potentially belong to the same class. However, even though we can control the degree of uncertainty in \(PFlag\) and \(NFlag\) through the parameter \(\alpha\), they may still propagate the noise in the embeddings space to the final feature vectors. 
Since the original lexical feature vectors are sparse, \(PFlag\) and \(NFlag\) may be awarded a high weight by the final classifier, and potentially cause more errors. To reduce the effect of these features, and also to utilize the association between the lexical features and their representation in the embeddings space, we constrain \(PFlag(t_i)\) and \(NFlag(t_i)\) to the region in the embeddings space in which \(t_i\) is located. Thus, we expect the two features also reflect the lexical similarity to some extent (in addition to storing information about the class label). The idea is illustrated in Figure~\ref{fig:2d-space-combined}(b).

Figure \ref{fig:2d-space-combined}(b) shows the same space that we discussed earlier with 3 hypothetical partitions, and two features for the examples mapped to each partition. Given a tweet \(t_i\) appearing in partition \(P_k\) feature \(PFlag_k(t_i)\) or \(NFlag_k(t_i)\) can be set. For instance, for the positive set of tweets which appear in partition \(P_2\), only the value of \(PFlag_2\) is set, and for the negative set of tweets which appear in partition \(P_3\), only the value of \(NFlag_3\) is set. We emphasize that this is different from the idea of clustering the embeddings space. The partitions in our case are used to represent the original posting text along with the class labels, since the tweets which are close in the embeddings space are likely to also share lexical content. On the other hand, we don't expect to have pure partitions due to the expected overlap in the vocabulary between the negative and positive classes. 

The number of partitions (\(K\)) can be tuned experimentally in the training phase. In general, we expect large partitions to improve recall, but to decrease precision. This is because larger partitions could result in a higher number of tweets to be mapped to the same pair of \(PFlag\) and \(NFlag\), which can potentially increase the number of detected positive cases, and also increase the chance of mislabeling the tweets. In the rest of the paper, we use the word \textit{we\_partitioning} to refer to the features proposed in this section.

\subsection{Distorting the Word Embeddings Space}
\label{subsec:we-mean}

In Section \ref{subsec:new-mean} we tried to partially address one of the drawbacks of directly using the word embeddings centroids, which is the loss of information about the constituent words. However, this fix does not resolve the inherent problem of using the centroids in the original word embeddings space. One approach to incorporate the information about individual terms is to integrate word importance into the computation of the tweet centroid vector. For instance, to reduce the effect of the less informative words on the centroid values, \cite{huang2012improving} suggests using IDF-weighting to compute the weighted average of the word vectors in the sentence representation context. In classification context, we propose to use information gain \cite{mitchell1997machine} weighting to compute the centroid vector to boost the impact of the words which are effective in the classification, effectively ``distorting'' the word embeddings space. More formally, we compute the new, ``distorted'' centroid of tweet \(t\) as:
\begin{displaymath}
	\vec{M_t} = \frac{\sum_{i=0}^{n}IG_i \times \vec{W_i}}{\sum_{i=0}^{n}IG_i}.
\end{displaymath}
where $\vec{M_t}$ is the weighted mean vector for tweet \(t\), $\vec{W_i}$ is the vector representation of word $W_i$ in tweet \(t\), and $n$ is the length of the tweet. $IG_i$ is the information gain of word $W_i$ in the training set, and is computed as:
\begin{displaymath}
	IG_i = Entr(D) - (\frac{|D_{w_i}|}{|D|} \times Entr(D_{w_i}) + \frac{|\overline{D_{w_i}}|}{|D|} \times Entr(\overline{D_{w_i}}))
\end{displaymath}
where \(D\) is the training set, \(Entr(D)\) is the entropy of \(D\) relative to our classification problem, \(|D|\) is the size of the training set, \(D_{w_i}\) is the subset of the training set for which \(W_i\) occurs, and \(\overline{D_{w_i}}\) is the subset of the training set for which \(W_i\) does not occur. For the words which do not appear in the training set, we estimate their information gain using the information gain of their closest word in the embeddings space, that do appear in the training set. 

\begin{figure}
\includegraphics[height=1.6in, width=3.3in]{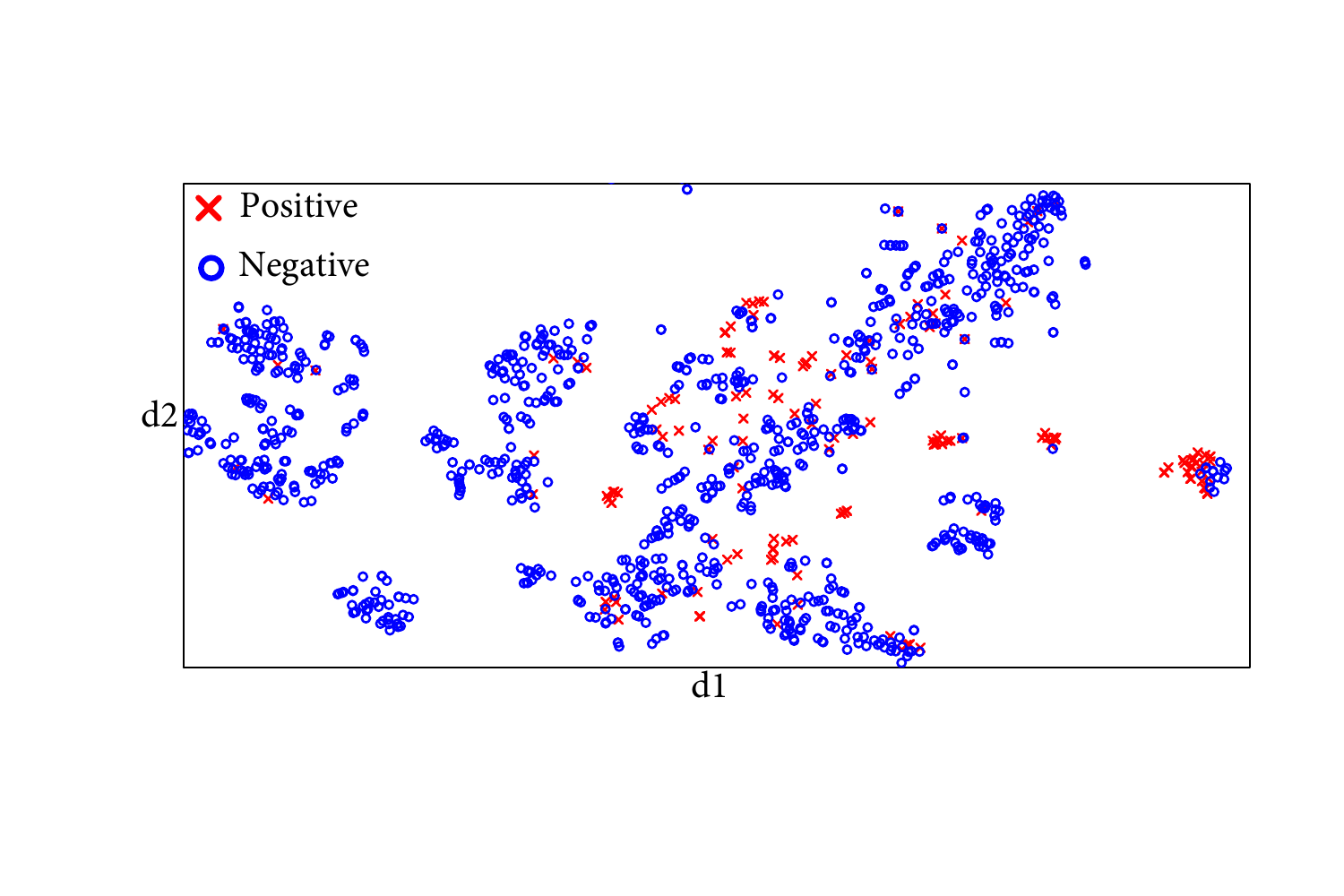}
\caption{The same set of tweet centroids reported in Figure \ref{fig:2d-space-combined}, after applying IG-weighting transformation.} \label{fig:2d-space-weighted}
\vspace{-0.5cm}
\end{figure}

Figure \ref{fig:2d-space-weighted} shows the same set of tweets from Figure~\ref{fig:2d-space-combined}, after applying ``IG-weighting''. The projection illustrates that in some cases, the transformation can successfully separate the tweets in different classes by mapping them to different regions of the word embeddings space.

To convert the new, weighed, centroids into features, we transform all the centroids using the information gain values extracted in the training set, and follow the model described in the previous sections to extract the centroid-based features for each tweet. The values of the parameter \(\alpha\) and the number of partitions \(K\), can be potentially different in the distorted word embeddings space, therefore, we call them \(\alpha_2\) and \(K_2\). In the rest of the paper, we use the term \textit{we\_distortion} to refer to the features introduced in this section.

\subsection{Representing the Posting Context} \label{subsec:context}
Previous studies in monitoring public health have shown that user posting history is a good indicator of his or her current state, \eg for depression detection \cite{de2013predicting}. We hypothesize that the users who post a message which includes a true personal health mention, might have already posted or will post a similar message. Although those messages may not necessarily contain the disease keywords, they may be semantically or lexically related to the current one. Therefore, we assume that true health-related postings will be somewhat consistent with the other, contemporaneous, posts by  the user. Of course, the actual effect of the health event on the user depends on a variety of factors, \eg the severity of the condition. To enable our model to capture these effects, in a way appropriate for each disease or condition of interest, we include a representation of the prior and subsequent posts by the user\footnote{Given the regular limitations in using social network API to retrieve the user postings, accessing the previous and next messages of the user might be problematic, specifically in the real-time large-scale applications.}. Therefore, we use the representation described in Sections~\ref{subsec:new-mean} and \ref{subsec:par-emb} to also represent the prior- and next- tweets of the user, and incorporate the resulting features into the final combined feature vector.  
In the subsequent sections, we use the terms \textit{context\_prev} and \textit{context\_next} to refer to the features extracted from the previous and next user messages respectively.


\section{\MODEL Classifier Implementation}
\label{sec:implementation}

So far, the proposed representation model provides a general approach for feature learning, and can be implemented with a number of different algorithms. We now describe the specific implementation to operationalize \MODEL into a classifier used for experiments in the rest of the paper. We emphasize that the implementation described below is just one (effective) way to operationalize the proposed model. 

\noindent\textbf{Lexical and syntactic features (Section~\ref{subsec:lex-syn})}: to parse and build the dependency tree for the tweet contents we used the parser introduced in \cite{kong2014dependency}. No stemming or stopword removal was performed\footnote{In our development experiments stemming and stopword removal was not helpful.}. To extract the frequent subtrees, we used the approach proposed in \cite{matsumoto2005sentiment}, with minimum support 10 and minimum tree size 2, as suggested in \cite{matsumoto2005sentiment}.

\noindent\textbf{Word embeddings implementation}: We experimented with multiple pre-trained word embeddings implementations. Specifically, we compared the {\em word2vec} word embeddings \cite{mikolov2013distributed} (with 300 dimensions), and the pretrained {\em GloVe} word embeddings \cite{pennington2014glove} (with 200 dimensions), specifically trained on Twitter data. We observed similar performance in both cases; for generality, we use the ``standard'' available {\em word2vec} word embeddings\footnote{Available at~\url{https://code.google.com/archive/p/word2vec/}.} for all of the reported experiments. Additional incremental improvements to our method may be achieved with further training of the word embeddings on domain-specific data, as done by some of the methods that we compare to in Section~\ref{sec:results}. 

\noindent\textbf{Detecting noisy regions in the word embeddings space (Section ~\ref{subsec:new-mean})}:  to detect the ``noisy'' regions in the word embeddings space, we implement the probabilistic mapping function \textit{Pr} by using the Mallet implementation \cite{mccallum2002mallet} of the multivariate logistic regression classifier with default settings. 

\noindent\textbf{Partitioning the word embeddings space (Section~\ref{subsec:par-emb})}:
to partition the word embeddings space into homogeneous regions we used the ELKI \cite{schubert2015framework} implementation of the {\em K}-means clustering algorithm. The value of K was chosen automatically for each task as described in Section \ref{subsec:train}. 
%

\noindent\textbf{Combining \MODEL features for health mention prediction}:
Finally, we combine the lexical, syntactic, word embedding-based, and context features described above into a joint model. For simplicity and interpretability, we used a logistic regression classifier, trained over the final feature vectors to label the tweets\footnote{We also experimented with an SVM classifier with linear kernel, and initially achieved slightly better results on development data. However, the improvement came at the cost of higher training time, therefore, we opted to stay with the simpler logistic regression model.}. Other classification algorithms, such as GBDT \cite{friedman2001greedy}, may potentially provide additional improvements by capturing non-linear relationships between the features, and may be explored in the future work.

For simplicity, the specific \MODEL implementation described above will be simply referenced as \MODEL for all of the reported experiments in the rest of the paper. 

\section{Experimental Setup} 
\label{sec:exper-setup}

We now describe the datasets that we used in the experiments, which include both an established benchmark dataset, and a new dataset created for the evaluation. Then we describe the baseline methods, and the experimental setup used for reporting and analyzing the results.

\subsection{Datasets} 
\label{subsec:datasets}

We used two datasets for training and evaluation. First, the prominent benchmark dataset introduced in reference~\cite{lamb2013separating}, focusing on identifying reports of influenza infection. This dataset, which we call {\em FLU2013} serves for calibration and benchmarking of our method and others, against a state-of-the-art method specifically designed for detecting Flu infection reports \cite{lamb2013separating}. To explore the scalability of the \PROBLEM detection of multiple diseases and conditions, we also created a new dataset, PHM2017, described below.\\ 

\noindent\textbf{FLU2013}:
this dataset was introduced in \cite{lamb2013separating}, and focused on separating \textit{awareness} of the disease from actual {\em infection} reports. Each tweet in the dataset was manually labeled into classes of flu awareness (negative) or flu report (positive). Since only Twitter IDs were distributed, the content of the tweets had to be retrieved for this study, which was done in winter 2017. At that time, there were 2,837 tweets still available to download, which is 63\% of the original dataset. There were 1,393 awareness (negative class) tweets, which account for 49\% of the dataset, and 1,444 report (positive class) tweets which account for 51\% of the  dataset.\\

\noindent\textbf{PHM2017}:
We also constructed a new dataset consisting of 7,192 English tweets across six diseases and conditions: Alzheimer's Disease, heart attack (any severity), Parkinson's disease, cancer (any type), Depression (any severity), and Stroke. We used the Twitter search API to retrieve the data using the colloquial disease names as search keywords, with the expectation of retrieving a high-recall, low precision dataset. After removing the re-tweets and replies, the tweets were manually annotated. The labels are:
\begin{itemize}[leftmargin=*]
\item \textit{self-mention}. The tweet contains a health mention with a health self-report of the Twitter account owner, \eg \textit{"However, I worked hard and ran for Tokyo Mayer Election Campaign in January through February, 2014, without publicizing the cancer."}
\item \textit{other-mention}. The tweet contains a health mention of a health report about someone other than the account owner, \eg \textit{"Designer with Parkinson's couldn't work then engineer invents bracelet + changes her world"}
\item \textit{awareness}. The tweet contains the disease name, but does not mention a specific person, \eg \textit{"A Month Before a Heart Attack, Your Body Will Warn You With These 8 Signals"}
\item \textit{non-health}. The tweet contains the disease name, but the tweet topic is not about health. \textit{"Now I can have cancer on my wall for all to see <3"}
\end{itemize}

\begin{table}
\centering
\begin{tabular}{|p{0.6in} |p{0.3in} |p{0.4in} |p{0.4in} |p{0.45in} |p{0.4in} |}\hline
\textbf{Topic} & 
{\footnotesize \textbf{Tweet count}} & 
{\footnotesize \textit{self-mention}} & 
{\footnotesize \textit{other-mention}} & 
{\footnotesize \textit{awareness}} & 
{\footnotesize \textit{non-health}} \\ \hline
Alzheimer & 1256 & 1\% & 17\% & 80\% & 2\% \\ \hline 
heart attack & 1219 & 4\% & 9\% & 17\% & 70\% \\ \hline 
Parkinson & 1040 & 2\% & 9\% & 65\% & 24\% \\ \hline 
cancer & 1242 & 3\% & 18\% & 62\% & 17\% \\ \hline 
depression & 1213 & 37\% & 3\% & 49\% & 11\% \\ \hline 
stroke & 1222 & 3\% & 11\% & 29\% & 57\% \\ \hline 
\end{tabular}
\caption{The distribution of tweets over topics and labels in \textbf{PHM2017} dataset.} 
\label{tbl-health}
\vspace{-0.5cm}
\end{table}

In our experiments, \textit{self-mention} and \textit{other-mention} labels are taken as positive class; and \textit{awareness} and \textit{non-health} labels are taken as negative class. To validate the labels, we engaged another annotator and randomly re-annotated 10\% of the tweets for each topic. Since we observed that the probability of having a disputed positive label is higher than having a disputed negative label, the 10\% re-labeling subset was drawn from the positive set. The re-annotation showed 85\% agreement between the annotators, which is acceptable for a challenging topic like health report. Table \ref{tbl-health} summarizes \textbf\textit{PHM2017} dataset. We observe that for each topic, a large portion of the tweets are in \textit{non-health} category, which shows that people tend to use these words in other contexts too--which confirms the previous findings \cite{yin2015scalable}. The statistics also show that, on average 19.5\% of the tweets in each topic are positive, which makes the classification task more challenging. Having both a balanced dataset (FLU2013) and an imbalanced dataset (PHM2017) helps us to evaluate our method in different settings.

To build the context features for \textbf\textit{PHM2017} dataset, we used Twitter API to download the user timelines. We were unable to build the context features for many of the tweets in the flu dataset, since in many cases either the timeline was unaccessible, or access to the user profile was restricted. Therefore, we report the results for FLU2013 without incorporating the context features, which, as we show, are helpful in PHM2017 dataset, and are expected to be available for many applications.

\subsection{Methods Compared}
\label{subsec:methods}

We implemented or adapted the following methods to compare \MODEL to both previously used methods for health classification, and to the latest classification methods based on deep neural networks that have shown promising performance for other tasks.
In Section \ref{sec:related}, we discussed that deep neural network classifiers need a large training set to reach their best performance; however, we included these state-of-the-art baselines to compare to the models which only rely on word embeddings.

\begin{itemize}[leftmargin=*]
\item \textit{ME+lex}. We used a logistic regression classifier (a.k.a Maximum Entropy classifier) trained over unigrams and bigrams.
\item \textit{ME+cen}. We used a logistic regression classifier trained over the text centroid representation of the tweets in the embeddings space.
\item \textit{ME+lex+emb}. We computed the text centroid representation of each tweet in the embeddings space, and combined the resulting vector with the unigrams and bigrams of the tweet. Then a logistic regression classifier was trained over the final vectors.
\item \textit{ME+lex+cen}. We added two features {\em PFlag} and {\em NFlag} to the corresponding vector of unigrams and bigrams of each tweet. Then we used the prediction of \textit{ME+cen} to set the values of {\em PFlag} to true if predicted positive, and {\em NFlag} to true if predicted negative. 
Finally, a logistic regression classifier was trained over the resulting vectors, to evaluate the contribution of our noise filtering method (Section \ref{subsec:new-mean}).
\item \textit{Rules}. Experiments in \cite{lamb2013separating} suggest that manually extracted templates and features are effective in detecting flu reports. We implemented the top six set of features reported in \cite{lamb2013separating}, and trained a logistic regression classifier over the resulting vectors. This model was used only in FLU2013 dataset.
\item \textit{CNN}. We used the convolutional neural network classifier introduced in \cite{kim2014convolutional}. We used the non-static variant, which can update the word vectors in the training. Using grid search, we tuned the number of convolution feature maps from values: \{50, 100, 150\}, and observed that the number of features highly depends on the training data, and thus was optimized automatically using grid search for each classification task. The rest of the hyperparameters were set to the suggested values\footnote{Available at~\url{https://github.com/harvardnlp/sent-conv-torch}}.
\item \textit{FastText}. We used the shallow neural network introduced in \cite{joulin2016bag}, known as FastText. This model represents the documents by taking the average over the individual word vectors, and can also update the vectors during the training. To tune the model we tried values: \{0.05, 0.1, 0.25, 0.5\} for learning rate, and values: \{2, 4\} for window size. We observed that the optimal value of the learning rate was not fixed, neither in FLU2013 nor in PHM2017. The value of window size was optimal at 4 in FLU2013, but was not fixed in PHM2017. The rest of the hyperparameters were set to the suggested values \cite{joulin2016bag}.
\item \textit{LSTM-GRNN}. We used the model proposed in \cite{tang2015document}, which is a two-step classifier. In the first step the model uses a long short-term memory neural network (LSTM) to produce the sentence representations, and in the second step, uses a gated recurrent neural network (GRNN) to encode the sentence relations in the document. 
Tweet dependency trees \cite{kong2014dependency} were used to detect sentence boundaries in order to produce the sentence representations.  To tune the model, we used the values: \{0.03, 0.3, 0.5\} for learning rate, and observed that it is optimal at 0.3 in FLU2013, but is not fixed in PHM2017. The rest of the hyperparameters were set to the suggested values in the original implementation~reference~\cite{tang2015document}.
\item \textit{\textbf{\MODEL}}: our method, described in Section~\ref{sec:model} and implemented as described in Section~\ref{sec:implementation}. 
\end{itemize}

\subsection{Training setup} 
\label{subsec:train}
To train and evaluate all of the methods in a fair and consistent way, 
we used the standard 10 fold Cross-Validation in FLU2013 dataset, and within each topic of PHM2017 dataset. The results reported in the next section are the averages over the test folds. To build the folds, we preserved the original distribution of the labels, and randomly assigned the tweets to each fold. Since the set of the positive tweets is small, we kept the folds fixed across all of the cross validation experiments, to ensure that all of the methods were trained and tested in identical train/validate/test folds and thus the results can be compared directly. 

We used grid search to tune the model hyper-parameters by maximizing the F1-measure in the target (positive) set. 
To tune the number of partitions \(K\) and \(K_2\) in the word embedding-based features of \MODEL (introduced in Sections \ref{subsec:par-emb} and \ref{subsec:we-mean}), we experimented with the values: \{3, 4, 5\}, and observed that their optimal values depended on the training data, and thus were chosen automatically for each task. To tune \(\alpha\) and \(\alpha_2\) (introduced in Sections \ref{subsec:new-mean} and \ref{subsec:we-mean}) we tried values: \{0.05, 0.15, 0.3\}, and observed the best performance for the value 0.05 in the FLU2013 dataset, and for the value 0.3, for all the topics, in PHM2017 dataset\footnote{For simplicity in the grid search we set \(\alpha=\alpha_2\).}.\\ 

\noindent\textbf{Evaluation Metrics}:
Since the proportion of the positive class in PHM2017 dataset was relatively low, the accuracy of all the models was high (on average 90\%), due primarily to accurately predicting the negative (majority) class--which is not as practically important as the target (positive) class (the true health mentions and the target of our study). Therefore, in the next section we report the {\bf F1-measure}, {\bf Precision}, and {\bf Recall} for  the positive class. 


\section{Results and Discussion} 
\label{sec:results}

We now report the experimental results. First, we report the main results in Section \ref{subsec:results}, followed by the discussion and feature analysis in Section~\ref{subsec:discuss}.

\subsection{Main Results} \label{subsec:results}

Table \ref{tbl-result-detail} reports F1-measure of all the models (described in Section \ref{subsec:methods}) across the topics in PHM2017 dataset. The experiments show that our model \MODEL outperforms all the baselines in the majority of the topics. The substantial difference in terms of F1-measure between \textit{ME+lex} and \MODEL models, shows that our model has successfully managed to learn the characteristics of the small set of the positive tweets, and to generalize better. Another observation is that model \textit{ME+lex+cen}, which uses lexical features alongside an output of a centroid-based classifier as an additional feature (see Section \ref{subsec:methods} for details), is performing relatively poorly. This validates our strategy described in \ref{subsec:new-mean} and \ref{subsec:par-emb}, and the need to detect and filter out the noisy regions in the word embeddings space. We can also see that \textit{CNN}, although with small amount of training data, is working surprisingly well. On the other hand, the complex \textit{LSTM+GRNN} model is outperformed on all the topics by our \MODEL classifier.

Table \ref{tbl-result-health} reports the average F1-measure, precision, and recall for all the models across the six topics in PHM2017 dataset. The results show that the main improvement of \MODEL comes from the higher recall, i.e., detecting additional true health mentions. Table \ref{tbl-result-health} also shows that the highest precision is achieved by the simple \textit{ME+lex} model, since this model only relies on the lexical features. On the other hand, \textit{LSTM+GRNN} has the lowest precision, and this can be attributed to the complex structure of the network which expects to be fine-tuned during the training.

\begin{table*}
\begin{tabular}{|p{0.7in} |p{0.5in} |p{0.6in} |p{0.5in} |p{0.4in} |p{0.5in} |p{0.4in} |}\hline
\textbf{Model} & 
{\footnotesize \textbf{Alzheimer's}} & 
{\footnotesize \textbf{Heart attack}} & 
{\footnotesize \textbf{Parkinson's}} & 
{\footnotesize \textbf{Cancer}} & 
{\footnotesize \textbf{Depression}} & 
{\footnotesize \textbf{Stroke}} \\ \hline 
\textit{ME+lex} & 0.701 & 0.399 & 0.468 & 0.533 & 0.722 & 0.610 \\ \hline 
\textit{ME+cen} & 0.704 & 0.327 & 0.383 & 0.587 & 0.727 & 0.453 \\ \hline 
\textit{ME+lex+emb} & 0.723 & 0.460 & 0.486 & 0.559 & 0.718 & 0.612 \\ \hline 
\textit{ME+lex+cen} & 0.720 & 0.415 & 0.464 & 0.628 & 0.737 & 0.601 \\ \hline 
\textit{LSTM-GRNN} & 0.725 & 0.482 & 0.617 & 0.624 & 0.676 & 0.564 \\ \hline 
\textit{FastText} & 0.769 & 0.491 & 0.540 & 0.605 & 0.741 & 0.633 \\ \hline 
\textit{CNN} & 0.767 & 0.554 & 0.653 & 0.622 & \textbf{0.768} & 0.676 \\ \hline 
\textit{\MODEL} & \textbf{0.800} & \textbf{0.571} & \textbf{0.672} & \textbf{0.670} & 0.758 & \textbf{0.698} \\ \hline 
\end{tabular}
\caption{F1-measure for the models across all the topics in PHM2017 dataset.} \label{tbl-result-detail}
\vspace{-0.5cm}
\end{table*}

\begin{table}[H]
\begin{tabular}{|p{0.9in} |p{0.5in} |p{0.5in} |p{0.5in} |}\hline
\textbf{Model} & 
{\footnotesize \textbf{F1}} & 
{\footnotesize \textbf{Precision}} & 
{\footnotesize \textbf{Recall}} \\ \hline
\textit{ME+lex} & 0.572 & \textbf{0.834} & 0.462 \\ \hline 
\textit{ME+cen} & 0.530 & 0.819 & 0.429 \\ \hline 
\textit{ME+lex+emb} & 0.593 & 0.833 & 0.483 \\ \hline 
\textit{ME+lex+cen} & 0.594 & 0.827 & 0.493 \\ \hline 
\textit{LSTM-GRNN} & 0.615 & 0.638 & 0.605 \\ \hline 
\textit{FastText} & 0.630 & 0.802 & 0.538 \\ \hline 
\textit{CNN} & 0.673 & 0.794 & 0.610 \\ \hline 
\textit{\MODEL} & \textbf{0.695} & 0.803 & \textbf{0.628} \\ \hline 
\end{tabular}
\caption{\underline{Average} F1-measure, precision, and recall in PHM2017 dataset.} \label{tbl-result-health}
\vspace{-0.5cm}
\end{table}

\begin{table}
\begin{tabular}{|p{0.9in} |p{0.5in} |p{0.5in} |p{0.5in} |}\hline
\textbf{Model} & 
{\footnotesize \textbf{F1}} & 
{\footnotesize \textbf{Precision}} & 
{\footnotesize \textbf{Recall}} \\ \hline
\textit{ME+lex} & 0.838 & 0.832 & 0.846 \\ \hline 
\textit{ME+cen} & 0.827 & 0.815 & 0.840 \\ \hline 
\textit{ME+lex+emb} & 0.843 & 0.837 & 0.850 \\ \hline 
\textit{ME+lex+cen} & 0.844 & 0.843 & 0.845 \\ \hline 
\textit{Rules} & 0.845 & 0.837 & 0.855 \\ \hline 
\textit{LSTM-GRNN} & 0.818 & 0.805 & 0.833 \\ \hline 
\textit{FastText} & 0.841 & 0.831 & 0.852 \\ \hline 
\textit{CNN} & 0.833 & \textbf{0.864} & 0.806 \\ \hline 
\textit{\MODEL} & \textbf{0.851} & 0.845 & \textbf{0.858} \\ \hline 
\end{tabular}
\caption{F1-measure, precision, and recall in FLU2013 dataset.} \label{tbl-result-flu}
\vspace{-0.5cm}
\end{table}

Table \ref{tbl-result-flu} reports F1-measure, precision, and recall of all the baselines in comparison to \MODEL in FLU2013 dataset. The results show that \MODEL outperforms all the baselines, even though there are considerable differences between PHM2017 and FLU2013 datasets (in terms of the proportion of the positive tweets). The results also show that \MODEL performs slightly better than the disease-specific \textit{Rules} classifier, implemented according to the descriptions in reference~\cite{lamb2013separating}. More detailed analysis revealed that the syntactic subtrees that we use in our model, to some extent, can also automatically capture the manually designed patterns reported in \cite{lamb2013separating}. It is also worth mentioning that, all the improvements of \MODEL model over the lexical baseline \textit{ME+lex} in both datasets are statistically significant using paired t-test at \textit{p$<$0.05}.

The comparison between the relative improvement of \MODEL in PHM2017 and FLU2013 datasets shows that our model performs significantly better in PHM2017 dataset. The improvement can be attributed to the inherent differences between these two datasets, and the fact that PHM2017 is highly imbalanced and FLU2013 is nearly balanced. We  discuss this issue further in the next section.


\subsection{Discussion} \label{subsec:discuss}
We now analyze the performance of \MODEL in more detail, focusing on the effects of the word embeddings partitioning, contribution of different features, and the ability of \MODEL to generalize from few positive examples in training. 

\begin{figure}
\includegraphics[height=2in, width=3in]{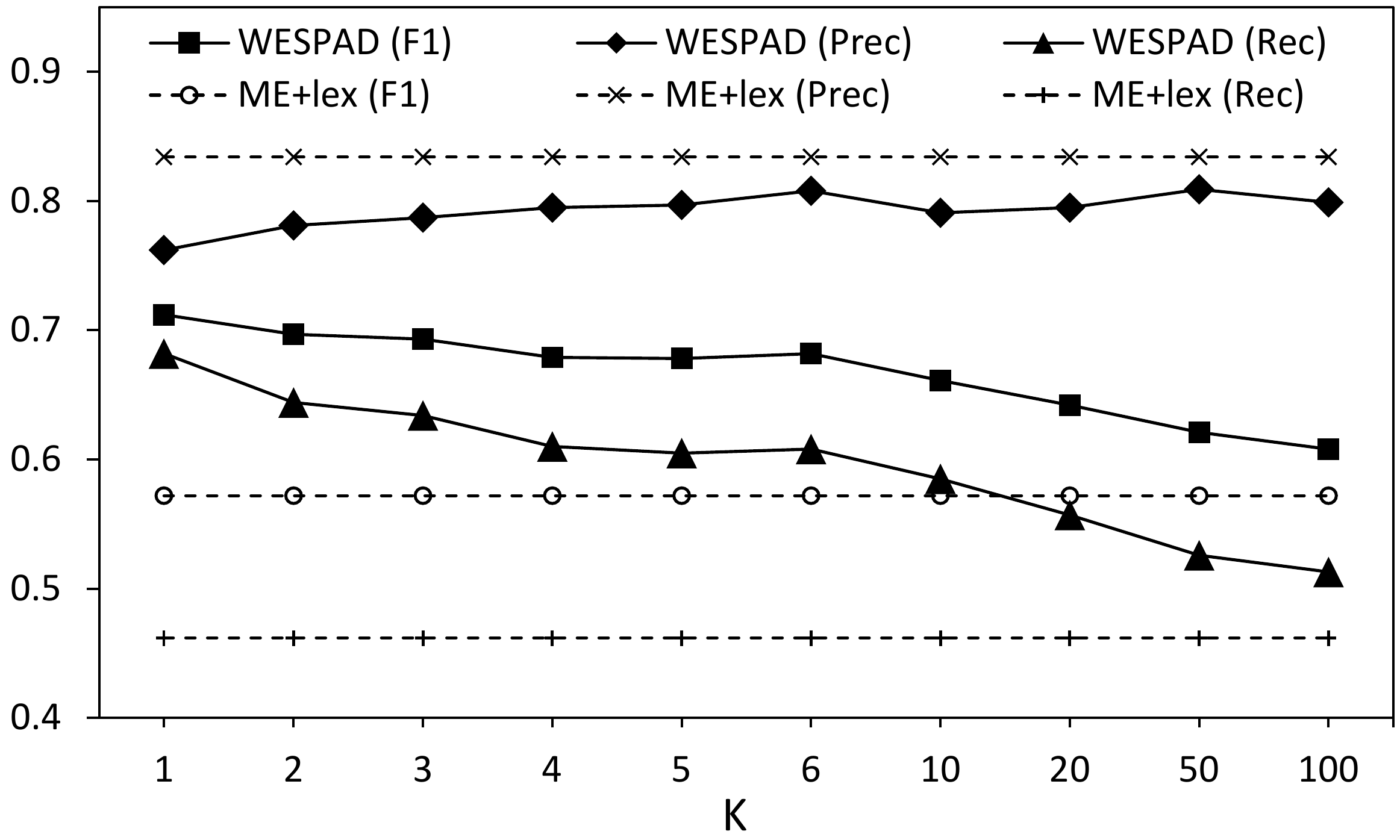}
\caption{Impact of the number of partitions \(K\) on \MODEL on F1-measure, precision, and recall (PHM2017 dataset).} \label{fig:kkk}
\vspace{-0.4cm}
\end{figure}

\noindent\textbf{Word embeddings partitions}: in Section~\ref{subsec:par-emb} we argued that large partitions can increase recall, and degrade precision. To support the argument, we fixed the values of \(\alpha\) and \(\alpha_2\); and experimented with different values for \(K\) (the number of the partitions in the regular embeddings space) and \(K_2\) (the number of the partitions in the distorted embeddings space). Figure \ref{fig:kkk} illustrates the result of this experiment. To be able to easier interpret the results, we also set \(K\) to be equal to \(K_2\). The experiment confirms that by decreasing the number of partitions (and thereby increasing the partition sizes), the {\em Recall} of \MODEL improves. However, this comes at the cost of degrading the {\em Precision} (specifically at \(K=1\)).

\noindent\textbf{Feature ablation}: Table \ref{tbl-ablation} reports the result of the ablation study on the features in \MODEL model in PHM2017 dataset. The experiment shows that \textit{we\_distortion} and \textit{we\_partitioning} feature sets have the highest impact, in terms of F1-measure. We also observe that, in terms of precision, \textit{we\_partitioning} performs better than \textit{we\_distortion}. One possible explanation is that due to the small size of the positive sets, IG-weighting may fail to accurately assign the weights to the word vectors, and thus, the tweet centroid is drifted.

\begin{table}
\begin{tabular}{|p{1.1in} |p{0.8in} |p{0.4in} |p{0.4in} |}\hline
\textbf{Feature set} & 
{\footnotesize \textbf{F1}} & 
{\footnotesize \textbf{Precision}} & 
{\footnotesize \textbf{Recall}} \\ \hline
\MODEL (all features) & 0.695 & 0.803 & 0.628 \\ \hline \hline
\textit{we\_distortion} & 0.643 (-7.4\%) & 0.804 & 0.554 \\ \hline 
\textit{we\_partitioning} & 0.652 (-6.1\%) & 0.788 & 0.578 \\ \hline 
\textit{context\_next} & 0.680 (-2.1\%) & 0.800 & 0.609 \\ \hline 
\textit{syn\_feats} & 0.682 (-1.8) & 0.800 & 0.613 \\ \hline 
\textit{context\_prev} & 0.686 (-1.2\%) & 0.801 & 0.616 \\ \hline 
\textit{context} & 0.687 (-1.1) & 0.795 & 0.620 \\ \hline 
\textit{lex\_feats} & 0.696 (+0.1) & 0.782 & 0.640 \\ \hline 
\end{tabular}
\caption{Feature ablation of \MODEL on PHM2017 dataset.} \label{tbl-ablation}
\vspace{-0.75cm}
\end{table}

\begin{figure}
\includegraphics[height=2in, width=3in]{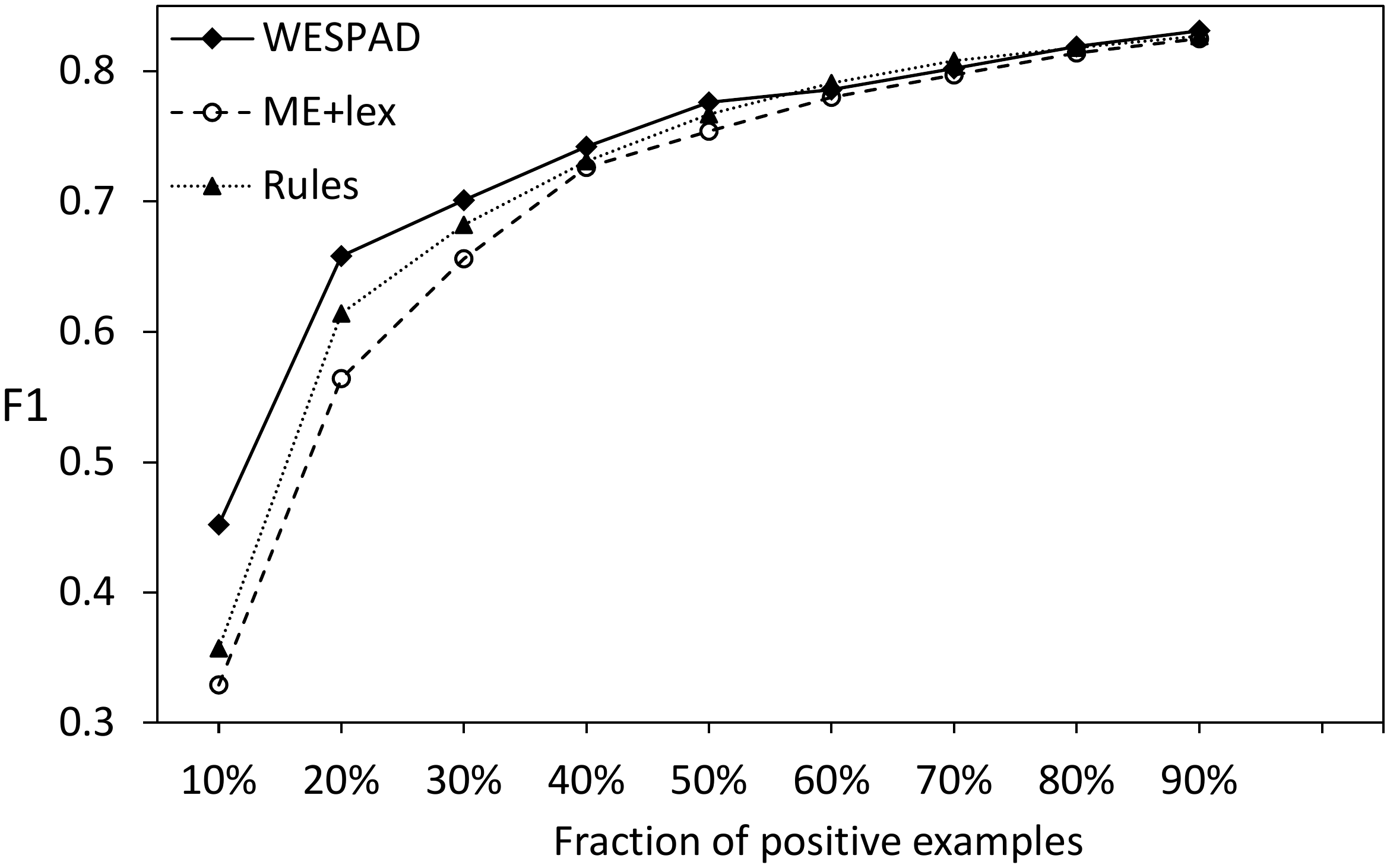}
\caption{F1 for \MODEL, \textit{Rules}, and \textit{ME+lex} trained on varying subsets of the positive examples (in FLU2013 dataset).} \label{fig:balance}
\vspace{-0.4cm}
\end{figure}

\noindent\textbf{Effect of the number of positive examples}: in Section~\ref{subsec:results} we observed that the relative improvement of \MODEL in PHM2017 dataset is considerably higher than its relative improvement in FLU2013 dataset. We argue that since FLU2013 dataset is nearly balanced, and also has a substantially larger set of positive tweets, simple models such as \textit{ME+lex} can perform relatively well.
To analyze the effect of the size of the training data, and specifically the availability of true positive examples, we varied the number of the positive examples in the training folds, by randomly sampling from 10\% to 90\% of the positive examples (and keeping all of the negative examples), and re-trained \MODEL, \textit{Rules}, and \textit{ME+lex} in the reduced training sets in FLU2013 dataset. Figure \ref{fig:balance} reports the values of the F1-measure for \textit{ME+lex}, \textit{Rules}, and \MODEL at varying fractions of the positive tweets used in the training data. The experiment shows that at smaller fractions of available positive tweets (10\%-30\%), \MODEL dramatically outperforms the \textit{ME+lex} baseline, demonstrating that \MODEL is able to generalize from fewer positive training examples. \MODEL also significantly outperforms \textit{Rules} at small fractions of positive tweets (10\%-20\%), signifying that the rule based models highly depend on their lexical based counterparts. We also observe that learning from just 20\% of the available positive examples, the F1-measure for \textit{ME+lex} model is 0.564, and for \MODEL model is 0.658. These F1 values are comparable to the F1 values that these models achieved in PHM2017 dataset, which also contains only 19\% of the positive class in the training and test data (on average, across the different disease topics).

In summary, our results show that WESPAD is able to outperform the state-of-the-art baselines for both datasets and under variety of settings, and even outperforms a disease-specific classifier in the prominent FLU2013 benchmark dataset. This is striking, as WESPAD does not require manual feature engineering, and can be trained with a relatively small number of (positive) training examples which makes WESPAD a valuable tool for extending health monitoring over social data to new diseases and conditions.



\section{Conclusions} 
\label{sec:conclusions}

We presented a new method, \MODEL, designed to detect personal health mentions in social data, such as Twitter posts. Unlike previously proposed methods for health classification, our method requires no manual feature engineering, and can be trained on relatively few positive examples of true health mentions. The improvements are due to a new approach to analyzing the representation of the examples in word embedding spaces, allowing \MODEL to discover a small number of effective features for classification. Furthermore, \MODEL can easily incorporate additional domain knowledge and can be extended to detect new diseases and conditions with relatively little effort. 

Our experimental evaluation compares \MODEL to a variety of previously proposed methods, including three state-of-the-art deep neural network approaches (LSTM, FasText, and CNN), on both an established benchmark dataset for detecting Flu infection reports, and a new PHM2017 dataset we created, with manual annotations of mentions for six different diseases and conditions. In the majority of the conditions, \MODEL exhibits superior overall performance.

By requiring a smaller number of training examples to achieve state-of-the-art performance, \MODEL can enable rapid development of domain-specific and robust text classifiers, which could in turn be valuable for tracking emerging diseases and conditions via social media.

\balance
{\footnotesize
\bibliographystyle{ACM-Reference-Format}
\bibliography{phm-bibliography} 
}
\end{document}